\documentclass{article}

\usepackage{microtype}
\usepackage{graphicx}
\usepackage{subfigure}
\usepackage{booktabs} % for professional tables
\usepackage{xfrac}

\usepackage{hyperref}

\usepackage[accepted]{icml2023}

\usepackage{amsmath}
\usepackage{amssymb}
\usepackage{mathtools}
\usepackage{amsthm}

\usepackage[capitalize,noabbrev]{cleveref}

\theoremstyle{plain}
\newtheorem{theorem}{Theorem}[section]

\newtheorem{lemma}[theorem]{Lemma}

\theoremstyle{definition}

\newtheorem{assumption}[theorem]{Assumption}
\newtheorem{condition}[theorem]{Condition}
\theoremstyle{remark}

\usepackage[textsize=tiny]{todonotes}

\usepackage{tikz}
\usetikzlibrary{patterns}
\usetikzlibrary{bayesnet}
\usetikzlibrary{arrows, positioning}
\usetikzlibrary{calc}
\usepackage{color}
\usetikzlibrary{backgrounds}
\usepackage{mathdots}
\usepackage{graphicx}
\usepackage{tabularx}
\usepackage{xfrac}
\definecolor{cambridgeblue}{rgb}{0.64, 0.76, 0.68}
\definecolor{colorCToTheta}{rgb}{1,0,0}
\definecolor{colorThetaToTheta}{rgb}{0.5, 0.5, 0.5}
\definecolor{colorThetaToX}{rgb}{0.4, 0.2, 0.8}
\definecolor{colorXToTheta}{rgb}{0.4, 0.8, 0.2}
\definecolor{forest}{rgb}{0.4, 0.8, 0.1}

\icmltitlerunning{Towards a Theoretical Understanding of CoT Prompting}

\begin{document}

\twocolumn[
\icmltitle{Why Can Large Language Models Generate Correct Chain-of-Thoughts?}



\icmlsetsymbol{equal}{*}

\begin{icmlauthorlist}
\icmlauthor{Rasul Tutunov}{equal,yyy}
\icmlauthor{Antoine Grosnit}{equal,yyy,compn}
\icmlauthor{Juliusz Ziomek}{yyy}
\icmlauthor{Jun Wang }{compa}
\icmlauthor{Haitham Bou-Ammar}{yyy,compa}
\end{icmlauthorlist}

\icmlaffiliation{yyy}{Huawei Noah's Ark}
%\icmlaffiliation{comp}{University of Oxford}
\icmlaffiliation{compa}{University College London}
% \icmlaffiliation{compo}{University of Oxford}
\icmlaffiliation{compn}{Technical University of Darmstadt}
%\icmlaffiliation{sch}{School of ZZZ, Institute of WWW, Location, Country}

\icmlcorrespondingauthor{Rasul Tutunov}{rasul.tutunov@huawei.com}
\icmlcorrespondingauthor{Antoine Grosnit}{antoine.grosnit2@huawei.com}

\icmlkeywords{Machine Learning, ICML}

\vskip 0.3in
]



\printAffiliationsAndNotice{\icmlEqualContribution} % otherwise use the standard text.

\begin{abstract}
This paper delves into the capabilities of large language models (LLMs), specifically focusing on advancing the theoretical comprehension of chain-of-thought prompting. We investigate how LLMs can be effectively induced to generate a coherent chain of thoughts. To achieve this, we introduce a two-level hierarchical graphical model tailored for natural language generation. Within this framework, we establish a \emph{compelling geometrical convergence rate} that gauges the likelihood of an LLM-generated chain of thoughts compared to those originating from the true language.
Our findings provide a theoretical justification for the ability of LLMs to produce the correct sequence of thoughts (potentially) explaining performance gains in tasks demanding reasoning skills.
\end{abstract}

\section{Introduction}
Since their inception \citep{OpenAI2023GPT4TR,touvron2023llama,jiang2023mistral,brown2020language,radford_language_2019}, large language models (LLMs) have revolutionised natural language processing \citep{zhang2023sentiment,wang2023chatgpt,ge2023openagi,lu2023sentiment,singhal2023towards,tan2023evaluation,guo2023images,huang2023enhancing,bohnet2022attributed,robinson2022leveraging} and seen widespread applications in a variety of fields, including but not limited to robotics \citep{singh2023progprompt,mai2023llm,vemprala2023chatgpt,huang2023visual,shah2023lm,liang2023code,ahn2022can}, medicine \citep{thirunavukarasu2023large,howard2023chatgpt,mbakwe2023chatgpt,gilbert2023large,nori2023capabilities,liu2023deid}, search \citep{kamalloo2023hagrid,vu2023freshllms}, content creation \citep{orenstrakh2023detecting,gmeiner2023dimensions}, code development \citep{liu2023your,wang2022code4struct,shen2023pangu,zheng2023codegeex,christopoulou2022pangucoder,chang2023learning}, customer support \citep{pesaru2023ai,markapudi2021new}. 
LLMs have amazingly demonstrated many emergent capabilities, e.g., chain-of-thought prompting, instruction fine-tuning, and in-context learning, when model sizes and training data grew large \citep{2023jiangTheory, wei2022emergent}.
Even more interesting is that LLMs exhibit new capabilities unobserved in small models or data scales, in contrast to simple scaling laws that convey generalisation within data distributions while only being trained on next-token prediction, as noted in \cite{2023jiangTheory,lu2023emergent,webb2023emergent,boiko2023emergent,noever2023numeracy,teehan2022emergent,2022WeiCoT}. 

Given those intriguing properties, machine learning researchers and practitioners have sought empirical and theoretical justifications that uncover the ``mystery'' behind those emergent behaviours. 
Empirically, works focused on understanding which training data properties may lead to emergent behaviour in LLMs and large-scale transformers. 
For instance, \citet{2022chanDataEmergent} demonstrates that in-context learning arises from dynamic clusters of items rather than uniformly distributed ones. Moreover, the authors of \cite{2022minRoleDemonstration} show that contrary to common belief, LLMs do not demand ground truth demonstration data to attain satisfactory performance in classification tasks. 
They instead discover label space coverage, input text distributions and the format of the input sequence as the main driving factors of in-context learning.
In another direction, the work in \citep{2022sanhMultitaskPrompt} notices that one can induce zero-shot generalisation via explicit multitask learning across prompts, while the authors in \citep{2022razeghiFSNumReason} study the importance of term frequencies during pretraining.    
\paragraph{{Theoretical Attempts of In-Context Learning:}} 
Apart from the above empirical studies, many authors have also attempted to provide a cohesive theoretical justification for such emergent behaviour from LLMs~\cite{2022xieIncontext, wies2023learnability, Hahn2023, 2023jiangTheory}. 
In their seminal work, \cite{2022xieIncontext} formalised in-context learning as latent concept discovery. 
They showed that LLMs learn to perform implicit Bayesian marginalisation during pretraining and that those models infer shared concepts at test time despite distribution mismatches between prompts and pretraining data. 
While providing a valid theoretical justification, their framework analyses the setup when the pretraining data distribution consists of a mixture of hidden Markov models (HMMs) and when an infinite limit of in-context examples are accessible.
The work in \citep{wies2023learnability} addressed the second limitation (i.e., infinite in-context examples), deriving finite sample complexity results under a PAC framework. 
Extending the results in \citep{2022xieIncontext} to general data distributions, the work in \citep{2023jiangTheory} explored sparsity properties present in joint distributions of languages and again demonstrated successful in-context learning when considering ambiguous and unambiguous latent language models.

\paragraph{{Chain-of-Thoughts Prompting:}} While in-context learning is an important emergent property of LLMs, improving performance by carefully designed input prompts is another intriguing property that led to remarkable successes, especially in mathematics and general reasoning tasks. In those reasoning domains, standard in-context prompting in which we ask the LLM to generate the final answer immediately often fails to yield the correct solutions.
To address such shortcomings, researchers proposed chain-of-thought (CoT) prompting that enabled LLMs to improve by triggering those models to output intermediate derivations \citep{2022WeiCoT, 2023fengExpressiveCoT}.
CoT prompting is generally performed in one of two ways: \textit{i)} by augmenting the input with specific phrases, e.g., ``Let us think step by step''~\cite{2022KojimaZeroShot}, or \textit{ii)} by providing a few-shot examples, see \citep{2022WeiCoT}.
Albeit enabling impressive empirical successes, CoT prompting remains a mystery, demanding better in-depth theoretical justification. 
Only recently did the authors of \citep{2023fengExpressiveCoT} attempt to provide a (partial) validation to uncover why CoTs can succeed. 
They tackled this problem from an expressivity perspective, revealing two fundamental results for a class of arithmetic problems:
\textit{i)} 
% bounded-depth transformers could only produce correct answers for arithmetic reasoning tasks if the model size grows super-polynomially compared to the input length, 
if we constrain a transformer-based LLM to output an answer directly, then the model depth should grow super-polynomially compared to the input length to cover any problem, while \textit{ii)} if we trigger the LLM to generate CoT, then a constant depth transformer is sufficient to solve any task from this class.

\paragraph{\underline{Contributions of This Study:}} The work above answers important (orthogonal) expressivity questions of auto-regressive transformer models but leaves unanswered \emph{how LLMs can be triggered to produce CoTs from this auto-regressive process}. In answering this question, we revert our attention to Bayesian inference and study the properties of LLMs as marginal approximators of natural language. 
As mentioned earlier, we are not the first to consider such a setup for LLMs \citep{2022xieIncontext,2023jiangTheory}. While similar in spirit to those works, directly applying the analysis of \citep{2022xieIncontext,2023jiangTheory} to justify CoTs is challenging and requires more realistic stochastic processes to account for the generation of a chain of thoughts. 
Consequently, we introduce a hierarchical graphical model with two-level latent variables denoting (unobserved) contexts - e.g., arithmetic operations or common-sense reasoning - and intentions. 
Importantly, we assume non-static latent intentions that evolve and generate the messages expressing the sequence of thoughts of the reasoning process for a true-to-practice setup. 
Notably, we assume that an intention $\boldsymbol{\theta}_i$ at step $i$ is conditionally generated on \emph{all previous messages and intentions and the context of the reasoning task} - a setup widely used in practice when producing CoTs from LLMs \citep{2023fengExpressiveCoT,2022WeiCoT}. 
This two-level hierarchical model that allows for the conditional evolution of latent intentions requires us to introduce new ambiguity definitions when deriving valid upper bounds on quantities of interest.
After presenting those, we further contribute by deriving a \emph{geometric upper bound} between the likelihood of the sequence of thoughts generated by an LLM versus those that would have been generated from the actual language. 
Section \ref{Sec:Theory}\ref{Sec:Theory} presents a formal exposure of our bounds. 
Now, we informally state our main result, providing intuition to the reader of what is to come: 
\begin{align*}
    |&p_{\texttt{LLM}}(\texttt{CoT}|\texttt{Inp}, \texttt{CoT-Examples(N)}) \\ 
    & \hspace{4em}- q_{\texttt{True}}(\texttt{CoT}|\texttt{Inp},\texttt{True-Context})| \leq \rho^{N},
\end{align*}
with $\rho < 1$ is a function of the language ambiguities.
% In other words, our results show that conditioned on an input and a set of N chain-of-thought examples, the likelihood of generating CoTs, without knowing the true context\footnote{Note that this is the standard setup under which LLMs operate when generating CoTs, i.e., from input and a set of CoT examples, the model tries to extrapolate a new set of thoughts that aid in solving the problem.}, i.e. pLLM(CoT|Inp,CoT-Examples(N))p_{\texttt{LLM}}(\texttt{CoT}|\texttt{Inp}, \texttt{CoT-Examples(N)}) is bounded by ρN\rho^{N} to the thoughts that would have been generated from the \emph{true} language which conditions on the \emph{true-context}, i.e., qTrue(CoT|Inp,True-Context)q_{\texttt{True}}(\texttt{CoT}|\texttt{Inp}, \texttt{True-Context}).
In other words, we consider the difference between the likelihood $p_{\texttt{LLM}}(\texttt{CoT}|\texttt{Inp}, \texttt{CoT-Examples(N)})$ of generating \texttt{CoT} without knowing the true context\footnote{Note that this is the standard setup under which LLMs operate to generate CoTs, i.e., from input and a set of CoT examples, the model tries to extrapolate a new sequence of thoughts that aid in solving the problem.}  but with a pre-prompt containing an input and a set of $N$ chain-of-thought examples, and the likelihood $q_{\texttt{True}}(\texttt{CoT}|\texttt{Inp}, \texttt{True-Context})$ of generating this same \texttt{CoT} from the \emph{true} language conditioned on the same input and the \emph{true-context}. Our results show that this difference is bounded by $\rho^{N}$, meaning that when triggered correctly, LLMs can generate the correct chain of thoughts that have been shown to improve performance on reasoning tasks.

%This paper provides new theoretical results on why LLMs can generate valid CoTs by introducing a realistic latent language generation process and analysing its properties. Under reasonable assumptions on the ambiguity of the natural language, we then derive a geometric upper bound between the likelihood of the sequence of thoughts generated by an LLM versus those that would have been generated from the actual language.  

\section{Chain-of-Thoughts Formulation}

\subsection{Chains of Thoughts in Natural Language.}
From a sensible natural language generation process viewpoint, a message is a sequence of tokens produced to convey information under a specific intention. 
Since the intention itself is not directly observable, it is a latent variable in the natural language generation process.
Moreover, as problem complexity increases, it becomes more natural  to arrive to the solution step by step  rather than  only providing the answer in a single message. 
A natural language generative model should therefore account for the generation of step by step solutions also known as chains of thought.
In a valid chain of thought prompt, each element of the chain is a message representing a thought, and the sequence of thoughts should be relevant and coherent to lead to the expected  answer~\cite{2023wangRelevant}. 
A message - and therefore its generative underlying intention - is relevant if it refers to some essential elements (precise names, figures, etc.) from the previous messages, and the sequence of thoughts is coherent if the $i^\text{th}$ step can be logically derived from the initial question and the previous thoughts.

\paragraph{An example of CoT:} To illustrate the notions of relevance and coherence 
introduced by \citet{2023wangRelevant}
, we analyse the first example of CoT provided in ~\cite {2022WeiCoT}, which we rewrite below using one colour per intermediary reasoning step.

\noindent\fbox{
\centering
    \parbox{.46\textwidth}{
        { Q: Roger has 5 tennis balls. He buys 2 more cans of
tennis balls. Each can have 3 tennis balls. How many
tennis balls does he have now?} \\

A: {\color{orange}Roger started with 5 balls.} {\color{cyan} 2 cans of 3 tennis balls
each is 6 tennis balls.} {\color{magenta} 5 + 6 = 11.} The answer is 11.
    }
}

As we show in Table~\ref{tab:CoTexample}, each message consists of a coherent and relevant step, which leads to the final answer. For instance, the last message of the reasoning, ``{\color{magenta} 5 + 6 = 11}", proceeds from a coherent thought consisting in adding the number of balls Roger started with (in the message, the relevant number ``5" is retrieved from the question) and the number of balls Roger acquired (retrieving the relevant number ``6" from the previous reasoning step) in order to arrive to the total number of tennis balls Roger has now.

\begin{table*}
\caption{Analysis of a chain of thoughts in terms of relevance and coherence.}
    \begin{tabularx}{\textwidth} { 
   >{\raggedright\arraybackslash}X 
  | >{\centering\arraybackslash}X 
  | >{\centering\arraybackslash}X  }
         Message&  Relevance& Coherence\\
         \hline
         {\color{orange} Roger started with 5 balls.}&  Refers to the ``5 tennis balls'' from the question.& Counts of the initial number of balls.\\
         \hline
         {\color{cyan}2 cans of 3 tennis balls each is 6 tennis balls.}&  Refers to "2 cans" and ``3 tennis balls'' from the question.& Computes the number of balls bought on top of the ones counted before.\\
         \hline
         {\color{magenta} 5 + 6 = 11} &  Refers to ``5'' and ``6'' from the two previous messages.& Uses previous steps counting independent sets of balls to get the total number of balls.
    \label{tab:CoTexample}
    \end{tabularx}
\end{table*}

\begin{figure}[h!]
    \centering
\resizebox{\linewidth}{!}{
\begin{tikzpicture}[->,>=stealth',shorten >=1pt,auto,node distance=1cm,
    main X/.style={,minimum size=1.2cm,thick,circle,draw,font=\sffamily\Large,fill=black!0},
    main theta/.style={,minimum size=1.2cm,thick,circle,draw,font=\sffamily\Large,fill=black!10},
    main C/.style={,minimum size=1.2cm,thick,circle,draw,font=\sffamily\Large,fill=black!30},
    ]

    \node[main theta] (theta0) {$\boldsymbol{\theta}_0$};
    \node[main theta] (theta1) [right=1.5cm of theta0]{$\boldsymbol{\theta}_1$};
    \node[main theta] (theta2) [right=1.5cm of theta1]{$\boldsymbol{\theta}_2$};
    \node[main theta] (thetam) [right=1.5cm of theta2]{$\boldsymbol{\theta}_M$};
    \node[main X] (X0) [below=1.5cm of theta0]{$\boldsymbol{x}_0$};
    \node[main X] (X1) [below=1.5cm of theta1]{$\boldsymbol{x}_1$};
    \node[main X] (X2) [below=1.5cm of theta2]{$\boldsymbol{x}_2$};
    \node[main X] (Xm) [below=1.5cm of thetam]{$\boldsymbol{x}_M$};
    \node[main C] (C) [above=1.6 of  $(theta0)!0.5!(thetam)$]{$\boldsymbol{c}$};
    % \node[main node] (C) [above=3.5 of  theta0]{$C$};
    \node[auto=false, above=.8 of theta2, text=colorCToTheta] { $\dots$ };
    % \node[auto=false, right=.6 of theta2, text=colorThetaToX] (interTheta) {$\dots$};
    % \node[auto=false, below=1 of interTheta, text=colorThetaToX] { $\dots$ };
    \node[auto=false] at ($(X2)!0.5!(Xm)$) {$\dots$};
    \node[auto=false, text=colorThetaToTheta] at ($(theta2)!0.5!(thetam)$) (interTheta) {$\dots$};

    \path[>=triangle 45, color=colorThetaToX] (theta0) edge (X0);
    \path[>=triangle 45, color=colorThetaToX] (theta1) edge (X1);
    \path[>=triangle 45, color=colorThetaToX] (theta2) edge (X2);
    \path[>=triangle 45, color=colorThetaToX] (thetam) edge (Xm);

    \draw[>=triangle 45, ->, solid, in=50, out=-170, color=colorCToTheta] (C) to (theta0);
    % \draw[>=triangle 45, ->, solid, in=90, out=-120] (C) to (theta1);
    % \draw[>=triangle 45, ->, solid, in=90, out=-60] (C) to (theta2);
    \draw[>=triangle 45, ->, solid, in=130, out=-10, color=colorCToTheta] (C) to (thetam);

    % \draw[>=triangle 45] (C) to (theta0);
    \draw[>=triangle 45, color=colorCToTheta] (C) to (theta1);
    \draw[>=triangle 45, color=colorCToTheta] (C) to (theta2);
    % \draw[>=triangle 45] (C) to (thetam);

    \draw[>=triangle 45, ->, solid, in=147, out=33, color=colorThetaToTheta] (theta0) to (thetam);
    \draw[>=triangle 45, ->, solid, in=165, out=25, color=colorThetaToTheta] (theta2) to (thetam);
    % \path[>=triangle 45] (theta2) -- node[auto=false]{$\ldots$} (thetam);
    \path[>=triangle 45, color=colorThetaToTheta] (theta0) edge (theta1);
    \path[>=triangle 45, color=colorThetaToTheta] (theta1) edge (theta2);
    \draw[>=triangle 45, ->, solid, in=140, out=23, color=colorThetaToTheta] (theta0) to (theta2);
    \draw[>=triangle 45, ->, solid, in=155, out=30, color=colorThetaToTheta] (theta1) to (thetam);
    \path[>=triangle 45, color=colorThetaToTheta] (theta2) edge (interTheta);
    \path[>=triangle 45, color=colorThetaToTheta] (interTheta) edge (thetam);

    % \draw[>=triangle 45, ->, solid, in=-160, out=80] (X0) to (theta1);
    % \draw[>=triangle 45, ->, solid, in=-160, out=70] (X0) to (theta2);
    % \draw[>=triangle 45, ->, solid, in=-140, out=80] (X1) to (theta2);
    % \path (X2) -- node[auto=false, rotate=54.4]{$\dots$} (thetam);
    % \draw[>=triangle 45, ->, solid, in=-165, out=55] (X0) to (thetam);
    % \draw[>=triangle 45, ->, solid, in=-150, out=55] (X1) to (thetam);
    % \draw[>=triangle 45, ->, solid, in=-135, out=75] (X2) to (thetam);

    \draw[>=triangle 45, color=colorXToTheta] (X0) to (theta1);
    \draw[>=triangle 45, color=colorXToTheta] (X0) to (theta2);
    \draw[>=triangle 45, color=colorXToTheta] (X1) to (theta2);
    \draw[>=triangle 45, color=colorXToTheta] (X0) to (thetam);
    \draw[>=triangle 45, color=colorXToTheta] (X1) to (thetam);
    \draw[>=triangle 45, color=colorXToTheta] (X2) to (thetam);
    
    % \path (X2) -- node[auto=false, rotate=54.4]{…\dots} (thetam);
    % \node[auto=false, rotate=45, above=2.5 of  (X2)!0.5!(Xm)(X2)!0.5!(Xm)]{…\dots};

\end{tikzpicture}
}
\caption{Probabilistic graphical model of natural language text generation that is compatible with the generation of chains of thoughts. $\boldsymbol{c}$ is a context, $(\boldsymbol{\theta}_i)_{0\leq i \leq M}$ is a sequence of intentions, and $(\boldsymbol{x}_i)_{0\leq i \leq M}$ is the sequence of messages corresponding to the formulated thoughts. 
The generation ends when the stop token is output $x_M= ``\langle  \text{END} \rangle$".
}
\label{fig:graph_model}
\end{figure}
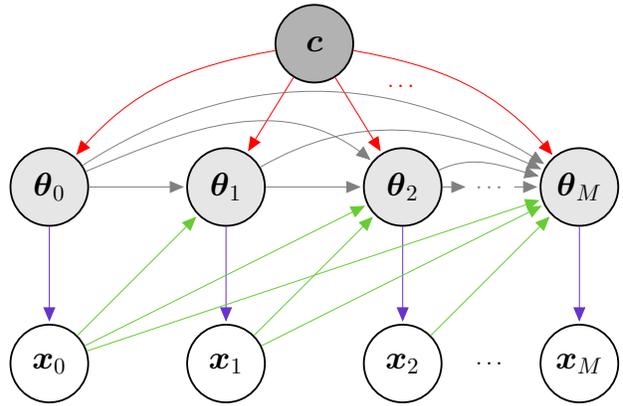

\paragraph{Probabilistic Graphical Model:}
Building from these observations and from previous  natural language latent models~\cite{2023jiangTheory,2022xieIncontext}, we propose in Figure~\ref{fig:graph_model} a new hierarchical latent probabilistic graphical model for natural language generation that can account for the generation of the chains of thought we find in corpora. 
The process starts by sampling a context $\boldsymbol{c}$ from a prior distribution $q(\boldsymbol{c})$ over the finite space of contexts $\boldsymbol{C}$. 
The abstract notion of context corresponds to an unobserved variable which can induce a specific type of logic (e.g. generating code in some language, doing arithmetic operations, imitating commonsense reasoning, etc.\footnote{These examples serve to give an idea of how to grasp this notion of context, but in reality, it is - as intentions - not observed and not as straightforward as what our examples show.}).
The initial intention $\boldsymbol{\theta}_0$ is generated based on the context $\boldsymbol{c}$, and the first message $\boldsymbol{x}_0$, which can be the description of a problem to solve, is sampled from a conditional distribution $q(\boldsymbol{x}_0|\boldsymbol{\theta}_0)$ such that the message conveys the information encapsulated in the intention.
At subsequent steps, the intention $\boldsymbol{\theta}_{i+1}$ is generated conditionally on all the previous messages $(\boldsymbol{x}_j)_{0\leq j \leq i}$ and intentions $(\boldsymbol{\theta}_j)_{0\leq j \leq i}$ to allow relevance, and on the overall context $\boldsymbol{c}$ responsible for the coherence of the chain.
At the lowest level, each message $\boldsymbol{x}_i$ is sampled conditioned only on its underlying intention $\boldsymbol{\theta}_i$.
Moreover, we assume the existence of a terminal intention $\boldsymbol{\theta}_\text{END}$ to which all sequences of messages converge, and such that $q(``\langle \text{END}\rangle"|\boldsymbol{\theta}_\text{END}) = 1$, where $``\langle  \text{END}\rangle"$ is the stop token. Therefore, we have the following probabilistic process that can generate chains of messages of variable lengths:
\begin{align*}
    & \boldsymbol{c} \sim q(\boldsymbol{c}), \  \boldsymbol{\theta}_0 \sim  q(\cdot|\boldsymbol{c}), \  \boldsymbol{x}_0 \sim  q(\cdot|\boldsymbol{\theta}_0) \\
     & \boldsymbol{\theta}_i \sim  q(\cdot|\boldsymbol{x}_{0:i-1}, \boldsymbol{\theta}_{0:i-1},\boldsymbol{c}), \  \boldsymbol{x}_i \sim  q(\cdot|\boldsymbol{\theta}_i) \quad \forall  i \geq 1 \\
     & \text{until}  \quad \boldsymbol{x}_{i} =  ``\langle  \text{END}\rangle"
\end{align*}

We assume that the corpora on which LLMs are trained come from i.i.d samples following the generative process described in Figure~\ref{fig:graph_model}. Notably, this does not imply that all the prompts from the pretraining datasets correspond to chains of thought since not all context variables in $\boldsymbol{C}$ induce coherent or relevant prompts. The context set should be broad enough to account for the diversity of the texts and speeches found in datasets.
% This assumption on the prompts distribution, along with the capacity of LLMs to serve as marginal approximators allow us to explain the success of CoT prompting with LLMs.

\subsection{Chain of Thoughts in LLMs}

\paragraph{LLMs as marginal approximators}

State-of-the-art LLMs~\cite{touvron2023llama, OpenAI2023GPT4TR} are large-scale autoregressive models based on transformer architectures~\cite{2017vaswaniTransformers} trained to perform text completion.
Concretely, an autoregressive LLM learns a conditional distribution over tokens $p_\texttt{LLM}(\cdot|\texttt{prompt})$ that should give more weights on the tokens that are the most likely to follow the given \texttt{prompt} according to the training data distribution.
At inference time, given any task description prompt $\boldsymbol{x}$, the LLM can generate an answer by  recursively predicting the sequence of next tokens from the learnt distribution $p_\texttt{LLM}$ conditioned on the concatenation of $\boldsymbol{x}$ and of the tokens sampled so far.
Interestingly, \citet{2023jiangTheory} showed that as model and training scales increase, LLM pretraining allows to capture the marginal distribution of the natural language, implying that for all sequences of messages $\boldsymbol{X} = (\boldsymbol{x}_0, \dots, \boldsymbol{x}_m)$ of at most $T$ tokens, $p_{\texttt{LLM}}(\boldsymbol{X})$ can be a good approximation of $q_\texttt{True}(\boldsymbol{X})$.  
This result relies on the fact that large enough transformer-based models can approximate any continuous sequence-to-sequence functions with arbitrary precision~\cite{2020YunUniApprox}, and in particular serve as approximators of $\boldsymbol{X} \mapsto q_\texttt{True}(\boldsymbol{X})$. 
Moreover, \citet{2023jiangTheory} established that if $p_{\texttt{LLM},n}$ maximises the empirical log-likelihood function $\operatorname{log}(p_{\texttt{LLM},n}(\boldsymbol{X}_1, \dots, \boldsymbol{X}_n))$ 
with $(\boldsymbol{X}_1, \dots, \boldsymbol{X}_n)$ being $n$ i.i.d. samples drawn from the natural language generative process, then the marginal distribution  $p_{\texttt{LLM},n}(\boldsymbol{X})$  converges to $q_\texttt{True}(\boldsymbol{X})$ for all sequences $\boldsymbol{X}$ as $n$ goes to infinity.  

In light of this theoretical result and considering that $p_{\texttt{LLM}}$ effectively matches the marginal distribution of $q_\texttt{True}$, we can analyse the empirical successes of the latest LLMs which benefited from a drastic increase in terms of model and training data sizes. 
Indeed, these enhancements allowed the emergence of capabilities unobserved with weaker models, such as the ability to trigger the generation of valid intermediate reasoning steps to solve tasks like arithmetic problems or symbolic reasoning~\cite{2022WeiCoT}. 
This feature is highly desirable for at least two reasons. Firstly, it allows the LLM to provide the correct solution to complex problems for which direct generation of the answer would fail, as documented in many previous works~\cite{2022WeiCoT, 2023wangRelevant, 2023yangLLMOptim,2022KojimaZeroShot, 2023suzgunBigbench}. 
Secondly, outputting a sequence of intermediary reasoning steps allows users to understand and validate the LLM's final answer and distinguish valid solutions from lucky guesses (or potentially indistinguishable unlucky ones).
This capacity can be elicited via proper conditioning of the LLM obtained with CoT prompting. 

\paragraph{Inferring context from CoT prompting}

To solve a task using an LLM generally consists of formulating the task in natural language to form an initial prompt $\boldsymbol{x}_0$ and to output an answer by recursively sampling one token at a time  from the learnt conditional distribution. 
Nevertheless, sampling the answer starting directly from $p_\texttt{LLM}(\cdot| \boldsymbol{x}_0)$ can lead to worse performance than starting from $\operatorname{warp}(\boldsymbol{x}_0)$, where $\operatorname{warp}$ is some strategy enriching the initial task description. 
A growing line of work is precisely dedicated to the design of prompt crafting strategies~\cite{fernando2023promptbreeder,chen2023evoprompting,guo2023connecting,chen2022program,cobbe2021training,lester2021power}, and endeavours to find warping functions $\operatorname{warp}$ that favour the sampling of the expected answer.
For reasoning tasks, \citet{2022WeiCoT} showed that few-shot CoT prompting improves the success rate by favouring the generation of valid CoTs. 

Formally, given a task $\boldsymbol{x}_0$ and its step-by-step solution $(\boldsymbol{x}_r)_{1\leq r \leq m}$ generated in natural language from the latent context $\boldsymbol{c}^*$, CoT prompting relies  on the warping $\operatorname{wrap}_{N-\texttt{shot}}$ to tighten the gap between $p_\texttt{LLM}((\boldsymbol{x}_r)_{1\leq r \leq m}|\operatorname{wrap}_{N-\texttt{shot}}(\boldsymbol{x}_0))$  and $q_\texttt{True}((\boldsymbol{x}_r)_{1\leq r \leq m}|\boldsymbol{x}_0, \boldsymbol{c}^*)$. Concretely, the $\operatorname{wrap}_{N-\texttt{shot}}$ strategy consists in adding to $\boldsymbol{x}_0$ a prompt made of $N$ examplar problems with their respective step-by-step answers $\boldsymbol{Z}_k = (\boldsymbol{z}_{k,r})_{0 \le r\le m_k}$ also coming from the same context $\boldsymbol{c}^*$, actually conditioning $p_\texttt{LLM}$  on $\boldsymbol{x}_0, (\boldsymbol{Z}_k)_{1 \leq k \leq N})$. 
Besides, $0$-shot-CoT prompting does not use any related CoT examples in its warping, but it elicits CoT output by adding an instruction prompt $\langle \texttt{INST} \rangle$  to $\boldsymbol{x}_0$ (e.g. ``Let's think step by step"~\cite{2022KojimaZeroShot}, ``Take a deep breath and work on this problem step-by-step"~\cite{2023yangLLMOptim}, etc.), which corresponds to having $\texttt{warp}_{0-\text{shot}}(\boldsymbol{x}_0) = (\boldsymbol{x}_0, \langle \texttt{INST} \rangle)$.
Since zero-shot-CoT does not perform on par with few-shot-CoT prompting~\cite{2022KojimaZeroShot}, \emph{we focus on the latter in this paper.} 

\paragraph{Natural Language Ambiguity:}
In the next section, we will show that the success of CoT prompting comes from the capacity of the LLMs to infer the true reasoning context from the provided series of examples  $(\boldsymbol{Z}_k)_{1 \le k \le N}$.
Nevertheless, the intrinsic ambiguity of natural language is a factor impeding the inference ability of LLMs since, by definition of ambiguity, it is not possible to fully identify the underlying intention or context of an ambiguous message or a sequence of messages. 
Formally, for a chain of messages
$(\boldsymbol{x}_i)_{0 \le i \le m}$ generated from our graphical model with latent context $\boldsymbol{c}^*$ and intentions $(\boldsymbol{\theta}_i^*)_{0 \le i \le m}$, we express the
ambiguity of the chain $\epsilon((\boldsymbol{x}_i)_{0 \le i \le m})$ as the complement of the likelihood of  the context $\boldsymbol{c}^*$ and intentions $(\boldsymbol{\theta}_i^*)_{0 \le i \le m}$ conditioned on $(\boldsymbol{x}_i)_{0 \le i \le m}$, i.e. we have
\begin{align*}
q_{\texttt{True}}(\boldsymbol{c}^*, (\boldsymbol{\theta}_i^*)_{0 \le i \le m}|(\boldsymbol{x}_i)_{0 \le i \le m})   = 1 -  \epsilon((\boldsymbol{x}_i)_{0 \le i \le m}).    
\end{align*}
This formulation extends the definition of ambiguity that \citet{2023jiangTheory} introduced to study in-context learning~\footnote{In \cite{2023jiangTheory}, the Section~$6$ discussing CoT only considers the case of unambiguous languages.}. 
For the in-context learning setup it is sufficient to define  the ambiguity of isolated prompts with respect to their unique underlying intentions $\epsilon(\boldsymbol{\theta}^*_{\boldsymbol{x}} | \boldsymbol{x})$ as the examples provided in in-context learning to the prompt are independent messages generated from the same intention.
On the other hand, in our setup the LLM is conditioned on sequences of prompts generated from a coherent chain of intentions guided by a hidden context, which requires to take into account the ambiguity of an entire sequence of messages with respect to all their underlying generative variables. 
Nevertheless, since  ambiguity goes against the core functional purpose of language (conveying information), it is reasonable to assume as in \cite{2023jiangTheory} that language evolved to decrease the overall level of ambiguity.

Next, we provide an upper bound on the difference between the likelihood of the reasoning steps under the true language $q_\texttt{True}(\cdot|\boldsymbol{x}_0, \boldsymbol{c}^*)$ and under the LLM conditioned via $k$-shot-CoT, and we express this bound as a product of ambiguity terms.

\section{LLMs can Produce Correct CoTs}\label{Sec:Theory}
This section presents our main results that shed light on how LLMs can effectively produce the correct sequence of thoughts, which, as mentioned earlier, has significantly enhanced their performance in various reasoning tasks. Before detailing our proof, we introduce and discuss an assumption on the prior distribution of the context $q(\boldsymbol{c})$. 

\begin{assumption}\label{assumption_uniform_contexts}
The prior distribution $q(\boldsymbol{c})$ associated with contexts $\boldsymbol{c} \in \boldsymbol{C}$ is uniform. %That is, for any c,c′∈C\boldsymbol{c}, \boldsymbol{c}^{'}\in \boldsymbol{C}:
    %\begin{align}\label{assumption_contx}
     %   q(\boldsymbol{c}) = q(\boldsymbol{c}^{'}).
    %\end{align}
\end{assumption}
The above assumption implies that natural language does not discriminate in some contexts over others. While this is valid when using large enough and well-balanced datasets, it can be violated in general, especially if data is imbalanced or skewed. In those cases, either more data can be collected, or self-supervised data augmentation can be adopted for skew correction \citep{yao2021self,reed2021selfaugment,liu2021self}. With that being said, in Section \ref{Sec:Relax}, we offer an alternative route that relaxes this assumption and can (potentially) lead us to a new bound for CoT generation with LLMs. 

\paragraph{\underline{Main Theorem:}}  Equipped with the above assumption, we now present our main result, which considers $N$-shot-CoT. 
In formalising our theorem, we follow the setup described in the previous section that we briefly summarise here for ease of exposure:\\
\noindent\rule{\columnwidth}{0.4pt}
\vspace{-2em}
\begin{center}   
\textbf{Problem Setup} 
\end{center}
\vspace{-1.2em}
\noindent\rule{\columnwidth}{0.4pt}
\vspace{-2em}
\begin{itemize}
    \setlength{\itemsep}{0pt}%
    \item The LLM is provided with $N$ \emph{varying length} chain of thought examples $\boldsymbol{Z}_{k}=(\boldsymbol{z}_{k,r})_{0\leq r\leq m_k}$ with $m_k$ denoting the length of the chain $\boldsymbol{Z}_k$ and each $\boldsymbol{z}_{k,r}$ being a sequence of tokens representing a thought or one reasoning step;
    \item Those examples are designed to aid the LLM in producing correct answers via CoT generation and thus, for all $k$, $\boldsymbol{Z}_{k}$ are generated with true intentions $(\boldsymbol{\theta}_{k,r}^{\star})_{0\leq r\leq m_k}$ and context $\boldsymbol{c}^{\star}$ in mind;
    \item Given $\boldsymbol{Z}_k$ and a task $\boldsymbol{x}_{0}$, the LLM then generates $(\boldsymbol{x}_r)_{1\leq r\leq m}$ messages. 
\end{itemize}
\vspace{-1.5em}
\noindent\rule{\columnwidth}{0.4pt}
At this stage, we can define the following: \textit{i)} the likelihood of the generated thoughts with our model, i.e. $p_{\texttt{LLM}}((\boldsymbol{x}_r)_{1\leq r\leq m}|\boldsymbol{x}_{0},\boldsymbol{Z}_{k})$, and \textit{ii)} the likelihood of the generated thoughts if we had known $\boldsymbol{c}^{\star}$, i.e. $q_{\texttt{True}}((\boldsymbol{x}_r)_{1\leq r\leq m}|\boldsymbol{x}_{0},\boldsymbol{c}^{\star})$. We then bound their differences in terms of the language's ambiguity to arrive at the following result:      

%how likely those thoughts are to be generated from the true language if we would have known the context c⋆\boldsymbol{c}^{\star}.       

\begin{theorem}\label{thmCoT}
Consider a collection of $N$ varying length chain-of-thought examples $\boldsymbol{Z}_k = (\boldsymbol{z}_{k,r})_{0 \le r\le m_k}$ generated from $(\boldsymbol{\theta}^*_{{k,r}})_{0\le r\le m_k}$ with a context $\boldsymbol{c}^* \sim q(\boldsymbol{c})$ that satisfies Assumption \ref{assumption_uniform_contexts}. Furthermore, let $\boldsymbol{x}_0 \sim q(\cdot|\boldsymbol{\theta}^{\star}_0)$ be the input message or task generated from $\boldsymbol{\theta}_0^{\star} \sim q(\cdot|\boldsymbol{c}^{\star})$.
Then, for any sequence of messages  $(\boldsymbol{x}_r)_{1\le r\le m}$ we have:
%Then, there are thresholds (m∗k∈N)1≤k≤N(m^*_k\in\mathbb{N})_{1 \le k\le N} such that for chain-of-thoughts examples sequences \boldsymbol{Z}_k\boldsymbol{Z}_k with lengths m_k\ge m^*_Km_k\ge m^*_K for any sequence of message prompt (\boldsymbol{x}_r)_{1\le r\le m}(\boldsymbol{x}_r)_{1\le r\le m} we have:
\begin{align}
    &|p_{\texttt{LLM}}((\boldsymbol{x}_r)_{1\le r\le m}|\boldsymbol{x}_0, (\boldsymbol{Z}_k)_{1\le k\le N}) \label{theorem_claim}\\ 
    & \qquad- q_{\texttt{True}}((\boldsymbol{x}_r)_{1\le r\le m}|\boldsymbol{x}_0,\boldsymbol{c}^*)| \le \eta \prod_{k=1}^N\frac{\boldsymbol{\epsilon}(\boldsymbol{Z}_k)}{1 - \boldsymbol{\epsilon}(\boldsymbol{Z}_k)}, \nonumber 
\end{align}
 with $\eta = 2\left( \sfrac{\epsilon(\boldsymbol{x}_0)}{1-\epsilon(\boldsymbol{x}_0)}\right)$ depending on the ambiguity of the input task. %
\end{theorem}
Informally, this result implies that a large language model prompted with chain-of-thought examples $(\boldsymbol{Z}_k)_{1\le k\le N}$ is capable of approximating the true natural language distribution equipped with a true context knowledge. 
The accuracy of this approximation is guided by ambiguity properties of provided examples $(\boldsymbol{\epsilon}(\boldsymbol{Z}_k))_{1\le k\le N}$ and input task $\epsilon(\boldsymbol{x}_0)$.
Notably, our result in Theorem \ref{thmCoT} holds for any sequence of messages $(\boldsymbol{x}_r)_{1\leq r \leq m}$. 
Consequently, this bound will also be valid for those messages generated by the LLM when prompted with $\boldsymbol{Z}_k$ and $\boldsymbol{x}_0$ in accordance with our ``Problem Setup''.   

\begin{proof}
   Appendix \ref{App:Proof} details the proof of Theorem  \ref{thmCoT}. Here, we provide a \emph{proof sketch} that presents the main steps needed to achieve our bound. Starting from $p_{\texttt{LLM}}((\boldsymbol{x}_{r})_{1\leq r \leq m} |\boldsymbol{x}_0, \boldsymbol{Z}_{k=1}^{N})$ and noticing that the LLMs' marginal distribution matches the true marginal, we can write:  
   \begin{align*}
       p_{\texttt{LLM}}((\boldsymbol{x}_{r})_{1\leq r \leq m} |\boldsymbol{x}_0, \boldsymbol{Z}_{k=1}^{N}) &= \\ 
       &\hspace{-1em}\frac{q_{\texttt{True}}((\boldsymbol{x}_{r})_{1\leq r \leq m}|\boldsymbol{x}_{0}, \boldsymbol{c}^{\star})+\mathcal{A}}{1+\mathcal{B}},
   \end{align*}
   where $\mathcal{A}$ and $\mathcal{B}$ are defined as:
   \begin{align*}
              \mathcal{A} &=\frac{\sum\limits_{\boldsymbol{c}\neq \boldsymbol{c}^*}q_{\texttt{True}}((\boldsymbol{x}_r)_{0\le r\le m}, \boldsymbol{Z}^N_{k=1}, \boldsymbol{c})}{q_{\texttt{True}}(\boldsymbol{Z}^N_{k=1}, \boldsymbol{x}_0, \boldsymbol{c}^*)} \quad  \\
        \mathcal{B} &= \sum\limits_{\boldsymbol{c}\neq \boldsymbol{c}^*}\frac{q_{\texttt{True}}(\boldsymbol{x}_0, \boldsymbol{Z}^N_{k=1}, \boldsymbol{c})}{q_{\texttt{True}}(\boldsymbol{Z}^N_{k=1}, \boldsymbol{x}_0, \boldsymbol{c}^*)}.
   \end{align*}
   We notice that although the input message $\boldsymbol{x}_0$ and the CoT examples $\boldsymbol{Z}^N_{k=1}$ share the same context $\boldsymbol{c}^*$, they are independent when conditioned on $\boldsymbol{c}^{\star}$. This observation allows us to establish the following bounds on $\mathcal{A}$ and $\mathcal{B}$: $$ \mathcal{A} \ \ \text{and} \ \ \mathcal{B}  \leq \frac{\boldsymbol{\epsilon}(\boldsymbol{x}_0)}{1 - \boldsymbol{\epsilon(\boldsymbol{x}_0)}}\prod_{k=1}^N\frac{\boldsymbol{\epsilon}(\boldsymbol{Z}_k)}{1 - \boldsymbol{\epsilon}(\boldsymbol{Z}_k)}.$$
  We arrive at the theorem's statement upon substituting those results in the absolute likelihood difference expression. 
\end{proof}

\section{What CoT Examples Will Work?}\label{Sec:Relax}
This section investigates sufficient requirements for CoT examples to guarantee vanishing bound in Equation~\ref{theorem_claim}. We start with the case when the CoT examples $\boldsymbol{Z}_{k}$ exhibit low ambiguity:
\begin{condition}\label{low ambig_assumption}
Imagine that the chain-of-thought examples $\boldsymbol{Z}_k = (\boldsymbol{z}_{k,r})_{0\le r\le m_k}$ generated from $(\boldsymbol{\theta}^*_{{k,r}})_{0\le r\le m_k}$ with a context $\boldsymbol{c}^* \sim q(\boldsymbol{c})$ have a bounded ambiguity measure such that:
\begin{align}\label{low_ambig_requirement}
    \boldsymbol{\epsilon}(\boldsymbol{Z}_k) = q_{\texttt{True}}(\boldsymbol{c}^*, (\boldsymbol{\theta}^*_{{k,r}})_{0\le r\le m_k}|(\boldsymbol{z}_{k,r})_{0\le r\le m_k}) \le \delta,
\end{align}
where $\delta \in [0,\frac{1}{2})$.
\end{condition}

This condition implies that the LLM model is prompted by carefully chosen CoT examples $\boldsymbol{Z}_k$ such that the true context $\boldsymbol{c}^*$ can be recovered from $\boldsymbol{Z}_k$ with reasonably high certainty (the probability the $\boldsymbol{c}^*$ is behind $\boldsymbol{Z}_k$ is strictly greater than one half). 
Provided with such CoT examples, one can establish a geometrical convergence rate for the bound in Equation \ref{theorem_claim} as the number of examples grows large:
\begin{align}\label{geometrical_convergence_bound}
    &|p_{\texttt{LLM}}((\boldsymbol{x}_r)_{1\le r\le m}|\boldsymbol{x}_0, (\boldsymbol{Z}_k)_{1\le k\le N}) \\\nonumber 
    & \hspace{5em}\qquad- q_{\texttt{True}}((\boldsymbol{x}_r)_{1\le r\le m}|\boldsymbol{x}_0,\boldsymbol{c}^*)| \le \eta \rho^\texttt{N},
\end{align}
where $\rho = \frac{\delta}{1 - \delta} \in [0,1)$.

Those examples described in Condition \ref{low ambig_assumption} should be carefully selected to guarantee low ambiguity requirements. In practice, however, it could be challenging to collect such chain-of-thought examples, as there is no rigorous procedure allowing us to measure ambiguity for a given sequence of thoughts. To remedy these strict requirements, we consider the following relaxed sufficient conditions:
\begin{condition}\label{length_conver_assumption}
    For the chain-of-thought examples $\boldsymbol{Z}_k = (\boldsymbol{z}_{k,r})_{0\le r\le m_k}$ generated from $(\boldsymbol{\theta}^*_{{k,r}})_{0\le r\le m_k}$ with a context $\boldsymbol{c}^* \sim q(\boldsymbol{c})$, the associated ambiguity measure $\epsilon((\boldsymbol{{z}_{k,r}})_{0\le r \le m_k})$ vanishes as length of sequence grows large:
\begin{align}\label{asympt_ambiguity}
        &\lim_{\ell\to\infty}\epsilon((\boldsymbol{{z}_{k,r}})_{0\le r \le \ell}) = 0.
    \end{align}
\end{condition}
Intuitively, the above condition implies that uncertainty over true context $\boldsymbol{c}^*$ and associated intentions $(\boldsymbol{\theta}^*_{k,r})_{0 \le r \le m_k}$ for a sequence of thoughts is diminishing when more of these thoughts are collected. 

The next lemma demonstrates that this asymptotic requirement is sufficient to guarantee low ambiguity measure for long enough CoT examples:
\begin{lemma}\label{lemma_on_lenth}
    Let us consider CoT examples $\{\boldsymbol{Z}_k = (\boldsymbol{z}_{k,r})_{0 \le r\le m_k}$ satisfying condition  \ref{length_conver_assumption}. Then, for any fixed $\delta\in [0,\frac{1}{2})$ there is a length threshold $m^*_{k,\delta}\in\mathbb{N}$, such that for any $m_k\ge m^*_{k,\delta}$:
    \begin{align*}
        &\epsilon(\boldsymbol{Z}_k) \le \delta
    \end{align*}
\end{lemma}
Appendix \ref{App:Proof_lemma} details the proof of Lemma  \ref{lemma_on_lenth}. This result implies in particular, that a geometrical convergence rate \ref{geometrical_convergence_bound} can also be established when the LLM model is prompted with CoT examples $\boldsymbol{Z}^N_{k=1}$ of sufficient length ($m_k\ge m^*_{k,\delta}$) satisfying Condition \ref{length_conver_assumption}. In contrast to the low ambiguity requirement, CoT examples with low asymptotic ambiguity can be more attainable. Indeed, being provided with high ambiguity CoT example $\boldsymbol{Z}_k$ satisfying Condition \ref{asympt_ambiguity}, one could break it to a longer sequence of thoughts $\boldsymbol{Z}^{'}_k$ consisting of more refined reasoning steps such that ambiguity of $\boldsymbol{Z}^{'}_k$ satisfies a sufficient threshold $\delta\in [0,\frac{1}{2})$.

%will  the behavior of the established bound under various assumptions on natural language statistical model qtrue(⋅)q_{\texttt{true}}(\cdot)  and CoT examples.  Although current large language models are trained on colossal datasets sampled from natural language distribution qtrue(⋅)q_{\texttt{true}}(\cdot), the diversity of possible contexts presented in these datasets can be skewed towards some contexts in favor of another. To investigate such scenarios we consider the following extension for a uniform prior distribution over contexts presented in Assumption ????????\ref{assumption_uniform_contexts}:

\paragraph{Non-Uniform Context Priors:} We stated our main result under the assumption that the prior distribution over the set of contexts is uniform.
Relaxing this assumption, we provide a similar bound on the discrepancy between the true and the predicted likelihood of a CoT.
To account for the potentially non-uniform distribution of contexts in training datasets, we introduce a skewness parameter $\gamma(\boldsymbol{c}^*)$, which we define as:
\begin{equation*}
    \gamma(\boldsymbol{c}^*) = 
    \underset{\boldsymbol{c}\in \boldsymbol{C}}{\operatorname{sup}} \ \frac{q_{\texttt{True}}(\boldsymbol{c}^*)}{q_{\texttt{True}}(\boldsymbol{c})}
\end{equation*}
Giving this measure of skewness, we establish a modified bound on the parameters $\mathcal{A}$ and $\mathcal{B}$: 
\begin{equation*}
    \mathcal{A} \ \ \text{and} \ \ \mathcal{B}  \leq \frac{\gamma^N(\boldsymbol{c}^*)\boldsymbol{\epsilon}(\boldsymbol{x}_0)}{1 - \boldsymbol{\epsilon(\boldsymbol{x}_0)}}\prod_{k=1}^N\frac{\boldsymbol{\epsilon}(\boldsymbol{Z}_k)}{1 - \boldsymbol{\epsilon}(\boldsymbol{Z}_k)}.
\end{equation*}

This, in turn, leading us to a modified bound of the following form:
\begin{align}
    &|p_{\texttt{LLM}}((\boldsymbol{x}_r)_{1\le r\le m}|\boldsymbol{x}_0, (\boldsymbol{Z}_k)_{1\le k\le N}) \label{theorem_claim_two}\\ 
    & \qquad- q_{\texttt{True}}((\boldsymbol{x}_r)_{1\le r\le m}|\boldsymbol{x}_0,\boldsymbol{c}^*)| \le \hat{\eta} \prod_{k=1}^N\frac{\boldsymbol{\epsilon}(\boldsymbol{Z}_k)}{1 - \boldsymbol{\epsilon}(\boldsymbol{Z}_k)}, \nonumber 
\end{align}
 with $\hat{\eta} =2 \gamma^N(\boldsymbol{c}^*) \frac{\boldsymbol{\epsilon}(\boldsymbol{x}_0)}{1 - \boldsymbol{\epsilon(\boldsymbol{x}_0)}}$ being a constant that depends on the ambiguity of the input task and skewness parameter $\gamma(\boldsymbol{c}^*)$.
 
%The above bound demonstrates that if skewness parameter $\gamma(\boldsymbol{c}^*)$ is large and ambiguity measures of CoT examples $(\epsilon(\boldsymbol{Z}_k))_{1\le k\le N}$  are not small enough to compensate for such value of $\gamma(\boldsymbol{c}^*)$  then the bound will become trivial. Intuitively, this implies that if some context $\boldsymbol{c}\in\mathcal{C}$ is severely underrepresented compared to the true context $\boldsymbol{c}^*$ in the natural language prior, then  the model is uncertain about the relation between 
 CoT examples $(\boldsymbol{Z}_k)_{1\le k\le N}$ and such context $\boldsymbol{c}$. Thus, to avoid such a situation, either the dataset of samples from $q_{\texttt{True}}(\cdot)$ should be diverse enough in terms of covering context set $\mathcal{C}$, and hence, guaranteeing small values for $\gamma(\boldsymbol{c}^*)$, or provided CoT examples should have small enough ambiguity, so the model with high certainty could guess the true context $\boldsymbol{c}^*$ from them.

\section{Conclusion}\label{Sec:Conclusion}

% \paragraph{Limitations and future directions:}
While the success of CoT prompting is well supported empirically, the emergent ability of LLMs to solve intricate tasks through intermediary steps is still not fully understood. We claim that our paper sheds light on one important aspect of the CoT mystery. 
While \citet{2023fengExpressiveCoT} studied the task-solving coverage of fixed-depth transformer-based LLMs, showing that the autoregressive generation of several messages permits significant expressivity gains, our paper gives insights into the role of the few-shot CoT prompting technique itself in the exploitation of this expressivity gain.
To do so, we have introduced a hierarchial latent language model  that  accounts for the generation of realistic chains of thought, i.e. of coherent and relevant series of messages. 
As emergent behaviours, such as LLM step-by-step reasoning ability, appear when the scale of architectures and training datasets increases, we followed the justification of~\cite{2023jiangTheory} to consider that LLM predictive distribution $p_\texttt{LLM}$ trained on next-token completion is a perfect approximator of the marginal distribution of the language model $q_\texttt{True}$. 
By framing the CoT-prompting as a wrapper of the task input that appends exemplary chains of thought, we showed that the LLM can leverage the examples to infer the underlying reasoning context. Therefore, conditioned on the task prompt and these examples, we derived an upper bound between the likelihood of the expected step-by-step answers under the, expressed as the product of ambiguity factors.
Finally, we thoroughly discussed the practical understanding of our theoretical bound, considering the introduction of different conditions on CoT ambiguity and providing a more general bound for non-uniform context priors. 
We hope that from these discussions, the success of CoT prompting appears less mysterious.

\paragraph{Future Work:} We plan to consider many interesting future research avenues. First, we wish to perform a rigorous experimental study to empirically validate our results in Section~\ref{Sec:Theory}. Second, we will investigate a more complete analysis than this paper by taking into account the separators in the actual prompts as part of our graphical model. Finally, we plan to generalise our analysis to cover recent discoveries in smart prompting, including but not limited to graph and tree of thoughts \cite{besta2023graph, yao2023tree}.

% we  experimental validations -- consider even more complete analysis taking into account the separators in the actual prompts as in HMM model. -- account for recent discoveries about CoT: invalid ones are as efficient as valid ones... Extend to GoTs ToTs and so on -- study the zero-shot CoT prompting.

%\section{Conclusion}
%\textbf{TO DO}

% \section*{Acknowledgements}

% The final camera-ready version can (and
% probably should) include acknowledgements. In this case, please
% place such acknowledgements in an unnumbered section at the
% end of the paper. Typically, this will include thanks to reviewers
% who gave useful comments, to colleagues who contributed to the ideas,
% and to funding agencies and corporate sponsors that provided financial

\bibliography{references}
\bibliographystyle{icml2023}

%%%%%%%%%%%%%%%%%%%%%%%%%%%%%%%%%%%%%%%%%%%%%%%%%%%%%%%%%%%%%%%%%%%%%%%%%%%%%%%
%%%%%%%%%%%%%%%%%%%%%%%%%%%%%%%%%%%%%%%%%%%%%%%%%%%%%%%%%%%%%%%%%%%%%%%%%%%%%%%
% APPENDIX
%%%%%%%%%%%%%%%%%%%%%%%%%%%%%%%%%%%%%%%%%%%%%%%%%%%%%%%%%%%%%%%%%%%%%%%%%%%%%%%
%%%%%%%%%%%%%%%%%%%%%%%%%%%%%%%%%%%%%%%%%%%%%%%%%%%%%%%%%%%%%%%%%%%%%%%%%%%%%%%
\newpage
\appendix
\onecolumn
\section{Proof of Theorem~\ref{thmCoT}} \label{App:Proof}

\begin{proof}
    We will give the proof for the general case of the prior distribution over contexts, and then provide the effect of Assumption \ref{assumption_uniform_contexts} on the obtained bound.
    Let us fix any sequence of messages $(\boldsymbol{x}_r)_{1\le r\le m}$ and consider the probability of a large language model to output $(\boldsymbol{x}_r)_{1\le r\le m}$ while being prompted with input message $\boldsymbol{x}_0$ and chain-of-thoughts examples $\boldsymbol{Z}^N_{k=1} = (\boldsymbol{Z}_k)_{1\le k\le N}$:
    \begin{align*}
        p_{\texttt{LLM}}((\boldsymbol{x}_r)_{1\le r\le m}|\boldsymbol{x}_0, \boldsymbol{Z}^N_{k=1}) &= \frac{q_{\texttt{True}}((\boldsymbol{x}_r)_{0\le r\le m}, \boldsymbol{Z}^N_{k=1})}{q_{\texttt{True}}(\boldsymbol{x}_0, \boldsymbol{Z}^N_{k=1})}\\\nonumber
        &=\frac{q_{\texttt{True}}((\boldsymbol{x}_r)_{0\le r\le m}, \boldsymbol{Z}^N_{k=1}, \boldsymbol{c}^*) + \sum\limits_{\boldsymbol{c}\neq \boldsymbol{c}^*}q_{\texttt{True}}((\boldsymbol{x}_r)_{0\le r\le m}, \boldsymbol{Z}^N_{k=1}, \boldsymbol{c})}{q_{\texttt{True}}(\boldsymbol{x}_0, \boldsymbol{Z}^N_{k=1}, \boldsymbol{c}^*) + \sum\limits_{\boldsymbol{c}\neq \boldsymbol{c}^*}q_{\texttt{True}}(\boldsymbol{x}_0, \boldsymbol{Z}^N_{k=1}, \boldsymbol{c})} 
    \end{align*}
    Notice,
    \begin{equation*}
        q_{\texttt{True}}((\boldsymbol{x}_r)_{0\le r\le m}, \boldsymbol{Z}^N_{k=1}, \boldsymbol{c}^*) =  q_{\texttt{True}}((\boldsymbol{x}_r)_{1\le r\le m}|\boldsymbol{x}_0, \boldsymbol{c}^*)q_{\texttt{True}}(\boldsymbol{Z}^N_{k=1}, \boldsymbol{x}_0, \boldsymbol{c}^*)
    \end{equation*}
    
    Hence, after diving both numerator and denominator on $q_{\texttt{True}}(\boldsymbol{Z}^N_{k=1}, \boldsymbol{x}_0, \boldsymbol{c}^*)$ gives:
    \begin{align}\label{Eq:llm_model}
        p_{\texttt{LLM}}((\boldsymbol{x}_r)_{1\le r\le m}|\boldsymbol{x}_0, \boldsymbol{Z}^N_{k=1}) &=         \frac{q_{\texttt{True}}((\boldsymbol{x}_r)_{1\le r\le m}|\boldsymbol{x}_0, \boldsymbol{c}^*) + \frac{\sum\limits_{\boldsymbol{c}\neq \boldsymbol{c}^*}q_{\texttt{True}}((\boldsymbol{x}_r)_{0\le r\le m}, \boldsymbol{Z}^N_{k=1}, \boldsymbol{c})}{q_{\texttt{True}}(\boldsymbol{Z}^N_{k=1}, \boldsymbol{x}_0, \boldsymbol{c}^*)}}{1 + \sum\limits_{\boldsymbol{c}\neq \boldsymbol{c}^*}\frac{q_{\texttt{True}}(\boldsymbol{x}_0, \boldsymbol{Z}^N_{k=1}, \boldsymbol{c})}{q_{\texttt{True}}(\boldsymbol{Z}^N_{k=1}, \boldsymbol{x}_0, \boldsymbol{c}^*)}}\\\nonumber
        &= \frac{q_{\texttt{True}}((\boldsymbol{x}_r)_{1\le r\le m}|\boldsymbol{x}_0, \boldsymbol{c}^*) + \mathcal{A}}{1 + \mathcal{B}}
    \end{align}
    where in the last step we use the notation:
    \begin{equation*}
        \mathcal{A} =\frac{\sum\limits_{\boldsymbol{c}\neq \boldsymbol{c}^*}q_{\texttt{True}}((\boldsymbol{x}_r)_{0\le r\le m}, \boldsymbol{Z}^N_{k=1}, \boldsymbol{c})}{q_{\texttt{True}}(\boldsymbol{Z}^N_{k=1}, \boldsymbol{x}_0, \boldsymbol{c}^*)} \quad \text{ and } \quad
        \mathcal{B} = \sum\limits_{\boldsymbol{c}\neq \boldsymbol{c}^*}\frac{q_{\texttt{True}}(\boldsymbol{x}_0, \boldsymbol{Z}^N_{k=1}, \boldsymbol{c})}{q_{\texttt{True}}(\boldsymbol{Z}^N_{k=1}, \boldsymbol{x}_0, \boldsymbol{c}^*)}
    \end{equation*}
    Let us study each of the above expressions separately. Using the fact that chain-of-thoughts examples $\boldsymbol{Z}^N_{k=1}$ are collected i.i.d and independent from messages $(\boldsymbol{x}_r)_{0\le r\le m}$ $\boldsymbol{x}_0$ we have (for any context $\boldsymbol{c}\in\boldsymbol{C}$):
    \begin{equation*}
        q_{\texttt{True}}((\boldsymbol{x}_r)_{0\le r\le m}, \boldsymbol{Z}^N_{k=1}, \boldsymbol{c}) = q_{\texttt{True}}((\boldsymbol{x}_r)_{1\le r\le m}| \boldsymbol{x}_0, \boldsymbol{c})\times q_{\texttt{True}}(\boldsymbol{x}_0|\boldsymbol{c})\prod_{k=1}^Nq_{\texttt{True}}(\boldsymbol{Z}_k|\boldsymbol{c})q_{\texttt{True}}(\boldsymbol{c})
    \end{equation*}
    and hence, we can write (using and definition of parameter $\gamma$):
    \begin{align*}
        \mathcal{A} &= \sum_{\boldsymbol{c}\neq \boldsymbol{c}^*}q_{\texttt{True}}((\boldsymbol{x}_r)_{0\le r\le m}| \boldsymbol{c}, \boldsymbol{x}_0) \frac{q_{\texttt{True}}(\boldsymbol{x}_0|\boldsymbol{c})q_{\texttt{True}}(\boldsymbol{c})}{q_{\texttt{True}}(\boldsymbol{x}_0|\boldsymbol{c}^*)q_{\texttt{True}}(\boldsymbol{c}^*)}\prod_{k=1}^N\left[\frac{q_{\texttt{True}}(\boldsymbol{Z}_k|\boldsymbol{c})q_{\texttt{True}}(\boldsymbol{c})}{q_{\texttt{True}}(\boldsymbol{Z}_k|\boldsymbol{c}^*)q_{\texttt{True}}(\boldsymbol{c}^*)}\right]\left[\frac{q_{\texttt{True}}(\boldsymbol{c}^*)}{q_{\texttt{True}}(\boldsymbol{c})}\right]^N  \\\nonumber
        &\le \gamma^N(\boldsymbol{c}^*)\sum_{\boldsymbol{c}\neq \boldsymbol{c}^*}\frac{q_{\texttt{True}}(\boldsymbol{x}_0|\boldsymbol{c})q_{\texttt{True}}(\boldsymbol{c})}{q_{\texttt{True}}(\boldsymbol{x}_0|\boldsymbol{c}^*)q_{\texttt{True}}(\boldsymbol{c}^*)}\sum_{\boldsymbol{c}\neq \boldsymbol{c}^*}\prod_{k=1}^N\left[\frac{q_{\texttt{True}}(\boldsymbol{Z}_k|\boldsymbol{c})q_{\texttt{True}}(\boldsymbol{c})}{q_{\texttt{True}}(\boldsymbol{Z}_k|\boldsymbol{c}^*)q_{\texttt{True}}(\boldsymbol{c}^*)}\right] \\\nonumber
        &\le \gamma^N(\boldsymbol{c}^*)\underbrace{\sum_{\boldsymbol{c}\neq \boldsymbol{c}^*}\frac{q_{\texttt{True}}(\boldsymbol{x}_0|\boldsymbol{c})q_{\texttt{True}}(\boldsymbol{c})}{q_{\texttt{True}}(\boldsymbol{x}_0|\boldsymbol{c}^*)q_{\texttt{True}}(\boldsymbol{c}^*)}}_{\mathcal{A}_1}\prod_{k=1}^N\underbrace{\left[\frac{\sum_{\boldsymbol{c}\neq \boldsymbol{c}^*}q_{\texttt{True}}(\boldsymbol{Z}_k|\boldsymbol{c})q_{\texttt{True}}(\boldsymbol{c})}{q_{\texttt{True}}(\boldsymbol{Z}_k|\boldsymbol{c}^*)q_{\texttt{True}}(\boldsymbol{c}^*)}\right]}_{\mathcal{A}_2}
    \end{align*}

    where we use the probability point mass function $q_{\texttt{True}}((\boldsymbol{x}_r)_{0\le r\le m}| \boldsymbol{c}, \boldsymbol{x}_0) \le 1$. Hence:
    \begin{align*}
        \mathcal{A}_1 = \sum_{\boldsymbol{c}\neq \boldsymbol{c}^*}\frac{q_{\texttt{True}}(\boldsymbol{x}_0, \boldsymbol{c})}{q_{\texttt{True}}(\boldsymbol{x}_0, \boldsymbol{c}^*)} = \frac{\sum_{\boldsymbol{c}\neq \boldsymbol{c}^*}\sum_{\boldsymbol{\theta}}q_{\texttt{True}}(\boldsymbol{x}_0, \boldsymbol{\theta},\boldsymbol{c})}{\sum_{\boldsymbol{\theta}}q_{\texttt{True}}(\boldsymbol{x}_0, \boldsymbol{\theta},\boldsymbol{c}^*)} &=\frac{q_{\texttt{True}}(\boldsymbol{x}_0) - \sum_{\boldsymbol{\theta}}q_{\texttt{True}}(\boldsymbol{\theta},\boldsymbol{c}^*| \boldsymbol{x}_0)q_{\texttt{True}}(\boldsymbol{x}_0)}{\sum_{\boldsymbol{\theta}}q_{\texttt{True}}(\boldsymbol{x}_0, \boldsymbol{\theta},\boldsymbol{c}^*)} \\
        %& = \frac{\sum_{\boldsymbol{c}\neq \boldsymbol{c}^*}\sum_{\boldsymbol{\theta}}q_{\texttt{True}}(\boldsymbol{\theta},\boldsymbol{x}_0,\boldsymbol{c})}{\sum_{\boldsymbol{\theta}}q_{\texttt{True}}(\boldsymbol{x}_0, \boldsymbol{\theta},\boldsymbol{c}^*)} \\
        %&=\frac{q_{\texttt{True}}(\boldsymbol{x}_0) - \sum_{\boldsymbol{\theta}}q_{\texttt{True}}(\boldsymbol{\theta},\boldsymbol{c}^*| \boldsymbol{x}_0)q_{\texttt{True}}(\boldsymbol{x}_0)}{\sum_{\boldsymbol{\theta}}q_{\texttt{True}}(\boldsymbol{x}_0, \boldsymbol{\theta},\boldsymbol{c}^*)} \\
        &\le \frac{q_{\texttt{True}}(\boldsymbol{x}_0) - q_{\texttt{True}}(\boldsymbol{\theta}_0,\boldsymbol{c}^*| \boldsymbol{x}_0)q_{\texttt{True}}(\boldsymbol{x}_0)}{q_{\texttt{True}}(\boldsymbol{x}_0, \boldsymbol{\theta}_0,\boldsymbol{c}^*)} \\
        &=\frac{q_{\texttt{True}}(\boldsymbol{x}_0) - q_{\texttt{True}}(\boldsymbol{\theta}_0,\boldsymbol{c}^*| \boldsymbol{x}_0)q_{\texttt{True}}(\boldsymbol{x}_0)}{q_{\texttt{True}}( \boldsymbol{\theta}_0,\boldsymbol{c}^*|\boldsymbol{x}_0)q_{\texttt{True}}(\boldsymbol{x}_0)} \\
        &=\frac{1 - q_{\texttt{True}}( \boldsymbol{\theta}_0,\boldsymbol{c}^*|\boldsymbol{x}_0)}{q_{\texttt{True}}( \boldsymbol{\theta}_0,\boldsymbol{c}^*|\boldsymbol{x}_0)} \\
        &= \frac{\boldsymbol{\epsilon}(\boldsymbol{x}_0)}{1 - \boldsymbol{\epsilon(\boldsymbol{x}_0)}}
    \end{align*}
    where in the last step we use the definition of ambiguity measure $\epsilon(\boldsymbol{x}_0) = q_{\texttt{True}}( \boldsymbol{\theta}_0,\boldsymbol{c}^*|\boldsymbol{x}_0)$. Next,
    \begin{align*}
        \mathcal{A}_2 = \frac{\sum_{\boldsymbol{c}\neq \boldsymbol{c}^*}q_{\texttt{True}}(\boldsymbol{Z}_k|\boldsymbol{c})q_{\texttt{True}}(\boldsymbol{c})}{q_{\texttt{True}}(\boldsymbol{Z}_k|\boldsymbol{c}^*)q_{\texttt{True}}(\boldsymbol{c}^*)} &= \frac{\sum_{\boldsymbol{c}\neq \boldsymbol{c}^*}\sum\limits_{(\boldsymbol{\theta}_{\boldsymbol{z}_{k,r}})_{0\le r\le m_k}}q_{\texttt{True}}(\boldsymbol{Z}_k, (\boldsymbol{\theta}_{\boldsymbol{z}_{k,r}})_{0\le r\le m_k}|\boldsymbol{c})q_{\texttt{True}}(\boldsymbol{c})}{\sum\limits_{(\boldsymbol{\theta}_{\boldsymbol{z}_{k,r}})_{0\le r\le m_k}}q_{\texttt{True}}(\boldsymbol{Z}_k, (\boldsymbol{\theta}_{\boldsymbol{z}_{k,r}})_{0\le r\le m_k}|\boldsymbol{c}^*)q_{\texttt{True}}(\boldsymbol{c}^*)} \\\nonumber
        &\le \frac{\sum_{\substack{\boldsymbol{c} \neq \boldsymbol{c}^*\\ (\boldsymbol{\theta}_{\boldsymbol{z}_{k,r}})_{0\le r\le m_k}}}q_{\texttt{True}}(\boldsymbol{Z}_k, (\boldsymbol{\theta}_{\boldsymbol{z}_{k,r}})_{0\le r\le m_k}|\boldsymbol{c})q_{\texttt{True}}(\boldsymbol{c})}{q_{\texttt{True}}(\boldsymbol{Z}_k, (\boldsymbol{\theta}^*_{\boldsymbol{z}_{k,r}})_{0\le r\le m_k}|\boldsymbol{c}^*)q_{\texttt{True}}(\boldsymbol{c}^*)}
    \end{align*}
    where $\boldsymbol{\theta}^*_{\boldsymbol{z}_{k,r}}$ is the intention behind message $\boldsymbol{z}_{k,r}$. Let us investigate the denominator term more carefully:
    \begin{align*}
        \sum_{\substack{\boldsymbol{c} \neq \boldsymbol{c}^*\\ (\boldsymbol{\theta}_{\boldsymbol{z}_{k,r}})_{0\le r\le m_k}}}q_{\texttt{True}}(\boldsymbol{Z}_k, (\boldsymbol{\theta}_{\boldsymbol{z}_{k,r}})_{0\le r\le m_k}|\boldsymbol{c})q_{\texttt{True}}(\boldsymbol{c}) &\le 
        \sum_{\substack{\boldsymbol{c} \neq \boldsymbol{c}^*\\ (\boldsymbol{\theta}_{\boldsymbol{z}_{k,r}})_{0\le r\le m_k}}}q_{\texttt{True}}(\boldsymbol{Z}_k, (\boldsymbol{\theta}_{\boldsymbol{z}_{k,r}})_{0\le r\le m_k}|\boldsymbol{c})q_{\texttt{True}}(\boldsymbol{c}) + \\\nonumber
        &\qquad\sum_{\substack{(\boldsymbol{\theta}_{\boldsymbol{z}_{k,r}})_{0\le r\le m_k} \\ \neq (\boldsymbol{\theta}^*_{\boldsymbol{z}_{k,r}})_{0\le r\le m_k}}}q_{\texttt{True}}(\boldsymbol{Z}_k, (\boldsymbol{\theta}_{\boldsymbol{z}_{k,r}})_{0\le r\le m_k}|\boldsymbol{c}^*)q_{\texttt{True}}(\boldsymbol{c}^*)\\\nonumber
        &\le \sum_{\substack{[\boldsymbol{c},(\boldsymbol{\theta}_{\boldsymbol{z}_{k,r}})_{0\le r\le m_k}] \\ \neq [\boldsymbol{c}^*,(\boldsymbol{\theta}^*_{\boldsymbol{z}_{k,r}})_{0\le r\le m_k}]}}q_{\texttt{True}}(\boldsymbol{Z}_k, (\boldsymbol{\theta}_{\boldsymbol{z}_{k,r}})_{0\le r\le m_k}|\boldsymbol{c})q_{\texttt{True}}(\boldsymbol{c})
        \\
        &=\sum_{\substack{[\boldsymbol{c},(\boldsymbol{\theta}_{\boldsymbol{z}_{k,r}})_{0\le r\le m_k}] \\ \neq [\boldsymbol{c}^*,(\boldsymbol{\theta}^*_{\boldsymbol{z}_{k,r}})_{0\le r\le m_k}]}}q_{\texttt{True}}(\boldsymbol{Z}_k, (\boldsymbol{\theta}_{\boldsymbol{z}_{k,r}})_{0\le r\le m_k},\boldsymbol{c}) \\\nonumber
        &= q_{\texttt{True}}(\boldsymbol{Z}_k) - q_{\texttt{True}}(\boldsymbol{Z}_k, (\boldsymbol{\theta}^*_{\boldsymbol{z}_{k,r}})_{0\le r\le m_k},\boldsymbol{c}^*) \\\nonumber
        &=q_{\texttt{True}}(\boldsymbol{Z}_k) - q_{\texttt{True}}( (\boldsymbol{\theta}^*_{\boldsymbol{z}_{k,r}})_{0\le r\le m_k},\boldsymbol{c}^*|\boldsymbol{Z}_k)q_{\texttt{True}}(\boldsymbol{Z}_k)
    \end{align*}
    Hence, we have:
    \begin{align*}
        \mathcal{A}_2 &\le \frac{q_{\texttt{True}}(\boldsymbol{Z}_k) - q_{\texttt{True}}( (\boldsymbol{\theta}^*_{\boldsymbol{z}_{k,r}})_{0\le r\le m_k},\boldsymbol{c}^*|\boldsymbol{Z}_k)q_{\texttt{True}}(\boldsymbol{Z}_k)}{q_{\texttt{True}}(\boldsymbol{Z}_k, (\boldsymbol{\theta}^*_{\boldsymbol{z}_{k,r}})_{0\le r\le m_k}|\boldsymbol{c}^*)q_{\texttt{True}}(\boldsymbol{c}^*)}\\\nonumber
        &=\frac{q_{\texttt{True}}(\boldsymbol{Z}_k)\left[1 - q_{\texttt{True}}( (\boldsymbol{\theta}^*_{\boldsymbol{z}_{k,r}})_{0\le r\le m_k},\boldsymbol{c}^*|\boldsymbol{Z}_k)\right]}{q_{\texttt{True}}(\boldsymbol{Z}_k)q_{\texttt{True}}( (\boldsymbol{\theta}_{\boldsymbol{z}_{k,r}})_{0\le r\le m_k},\boldsymbol{c}^*|\boldsymbol{Z}_k)} \\\nonumber
        &=\frac{1 - q_{\texttt{True}}( (\boldsymbol{\theta}_{\boldsymbol{z}_{k,r}})_{0\le r\le m_k},\boldsymbol{c}^*|\boldsymbol{Z}_k)}{q_{\texttt{True}}( (\boldsymbol{\theta}_{\boldsymbol{z}_{k,r}})_{0\le r\le m_k},\boldsymbol{c}^*|\boldsymbol{Z}_k)} \\ 
        &= \frac{\boldsymbol{\epsilon}(\boldsymbol{Z}_k)}{1 - \boldsymbol{\epsilon}(\boldsymbol{Z}_k)}
    \end{align*}
    where in the last step we use the definition of ambiguity measure for chain-of-thoughts $\boldsymbol{Z}_k$. Combining results for $\mathcal{A}_1$ and $\mathcal{A}_2$ gives:
    \begin{align*}
        &\mathcal{A} \le \frac{\gamma^N(\boldsymbol{c}^*)\boldsymbol{\epsilon}(\boldsymbol{x}_0)}{1 - \boldsymbol{\epsilon(\boldsymbol{x}_0)}}\prod_{k=1}^N\frac{\boldsymbol{\epsilon}(\boldsymbol{Z}_k)}{1 - \boldsymbol{\epsilon}(\boldsymbol{Z}_k)}
    \end{align*}
    Similarly, we establish the bound
    \begin{align*}
        \mathcal{B} \le \frac{\gamma^N(\boldsymbol{c}^*)\boldsymbol{\epsilon}(\boldsymbol{x}_0)}{1 - \boldsymbol{\epsilon(\boldsymbol{x}_0)}}\prod_{k=1}^N\frac{\boldsymbol{\epsilon}(\boldsymbol{Z}_k)}{1 - \boldsymbol{\epsilon}(\boldsymbol{Z}_k)}
    \end{align*}
    Hence, using the above  results we have:
    \begin{align*}
        |p_{\texttt{LLM}}((\boldsymbol{x}_r)_{1\le r\le m}|\boldsymbol{x}_0, \boldsymbol{Z}^N_{k=1}) - q_{\texttt{True}}((\boldsymbol{x}_r)_{1\le r\le m}|\boldsymbol{x}_0, \boldsymbol{c}^*)| 
        &=\frac{|\mathcal{A} - \mathcal{B}q_{\texttt{True}}((\boldsymbol{x}_r)_{1\le r\le m}|\boldsymbol{x}_0, \boldsymbol{c}^*)|}{1 + \mathcal{B}}  \\\nonumber
        &\le |\mathcal{A} + \mathcal{B}q_{\texttt{True}}((\boldsymbol{x}_r)_{1\le r\le m}|\boldsymbol{x}_0, \boldsymbol{c}^*)| \\
        &\le \mathcal{A} + \mathcal{B} \\\nonumber
        &\le a\prod_{k=1}^N\frac{\boldsymbol{\epsilon}(\boldsymbol{Z}_k)}{1 - \boldsymbol{\epsilon}(\boldsymbol{Z}_k)} 
    \end{align*}
    where $a = 2\frac{\gamma^N(\boldsymbol{c}^*)\boldsymbol{\epsilon}(\boldsymbol{x}_0)}{1 - \boldsymbol{\epsilon}(\boldsymbol{x}_0)}$. Hence, in case Assumption \ref{assumption_uniform_contexts} holds we have $\gamma(\boldsymbol{c}^*) = 1$ and $a = 2\frac{\boldsymbol{\epsilon}(\boldsymbol{x}_0)}{1 - \boldsymbol{\epsilon}(\boldsymbol{x}_0)}$
    
\end{proof}

\section{Proof of Lemma~\ref{lemma_on_lenth}} \label{App:Proof_lemma}
\begin{proof}

Let us fix $\delta\in [0, \frac{1}{2})$, then for CoT  $\boldsymbol{Z}_k = (\boldsymbol{z}_{k,r})_{0\le r \le m_k}$ satisfying Assumption \ref{length_conver_assumption} we have:
    \begin{align*}
        \lim_{m_k\to\infty}\epsilon(\boldsymbol{Z}_k) = 0
    \end{align*}
    then there exists  $m^*_{k,\delta}\in\mathbb{N}$ such that for any $m_k\ge m^*_{k,\delta}$ we have:
    \begin{align*}
        \epsilon(\boldsymbol{Z}_k) \le \delta
    \end{align*}
which finishes the proof of the Lemma.    
\end{proof}

%%%%%%%%%%%%%%%%%%%%%%%%%%%%%%%%%%%%%%%%%%%%%%%%%%%%%%%%%%%%%%%%%%%%%%%%%%%%%%%
%%%%%%%%%%%%%%%%%%%%%%%%%%%%%%%%%%%%%%%%%%%%%%%%%%%%%%%%%%%%%%%%%%%%%%%%%%%%%%%

\end{document}